\documentclass[11pt,a4paper]{article}
\usepackage[hyperref]{naaclhlt2019}
\usepackage{times}
\usepackage{latexsym}
\usepackage{graphicx}
\usepackage{amsmath}

\usepackage{paralist}
\usepackage{tikz}
\usetikzlibrary{shapes,arrows,positioning}
\usetikzlibrary{matrix}
\usepackage{etoolbox}

\usepackage[english]{babel}
\usepackage[T1]{fontenc}
\usepackage[utf8]{inputenc} 

\usepackage{url}

\usepackage{todonotes} 

\aclfinalcopy 

\title{Understanding Learning Dynamics Of Language Models with SVCCA}

\author{Naomi Saphra  \and Adam Lopez \\
{\tt n.saphra@ed.ac.uk} \phantom{\and} {\tt alopez@ed.ac.uk} \\
  Institute for Language, Cognition, and Computation \\
  University of Edinburgh \\}

\begin{document}
\maketitle

\begin{abstract}
    Research has shown that neural models implicitly encode linguistic features, but there has been no research showing \emph{how} these encodings arise as the models are trained. We present the first study on the learning dynamics of neural language models, using a simple and flexible analysis method called Singular Vector Canonical Correlation Analysis (SVCCA), which enables us to compare learned representations across time and across models, without the need to evaluate directly on annotated data. We probe the evolution of syntactic, semantic, and topic representations and find that part-of-speech is learned earlier than topic; that recurrent layers become more similar to those of a tagger during training; and embedding layers less similar. Our results and methods could inform better learning algorithms for NLP models, possibly to incorporate linguistic information more effectively.
\end{abstract}

\section{Introduction} \label{sec:intro}

Large neural networks have a notorious capacity to memorize training data  \citep{zhang2016understanding}, but their high accuracy on many NLP tasks shows that they nonetheless generalize. One apparent explanation for their performance is that they learn linguistic generalizations even without explicit supervision for those generalizations---for example, that subject and verb number agree in English \citep{linzen_assessing_2016}; that derivational suffixes attach to only specific parts of speech \citep{kementchedjhieva2018indicatements}; and that short segments of speech form natural clusters corresponding to phonemes \citep{alishahi2017encoding}. These studies tell us that neural models learn to implicitly represent linguistic categories and their interactions. But \textit{how} do they learn these representations?

One clue comes from the inspection of multi-layer models, which seem to encode lexical categories in lower layers, and more contextual categories in higher layers. For example,  \citet{blevins_deep_2018} found that a word’s part of speech (POS) is encoded by lower layers, and the POS of its syntactic parent is encoded by higher layers; while \citet{belinkov_evaluating_2018} found that POS is encoded by lower layers and semantic category is encoded by higher layers. More generally, the most useful layer for an arbitrary NLP task seems to depend on how ``high-level'' the task is \citep{peters2018deep}. Since we know that lower layers in a multi-layer model converge to their final representations more quickly than higher layers \citep{raghu_svcca:_2017}, it is likely that models learn local lexical categories like POS earlier than they learn higher-level linguistic categories like semantic class.

How and when do neural representations come to encode specific linguistic categories? Answers could explain why neural models work and help us improve learning algorithms. 
We investigate how representations of linguistic structure are learned over time in neural language models (LMs), which are central to NLP: on their own, they are used to produce contextual representations of words for many tasks \citep[e.g.][]{peters2018deep}; while \emph{conditional} LMs power machine translation, speech recognition, and dialogue systems. We use a simple and flexible method, Singular Vector Canonical Correlation Analysis \citep[SVCCA;][]{raghu_svcca:_2017}, which allows us to compare representations from our LM at each epoch of training with representations of other models trained to predict specific linguistic categories. We discover that lower layers initially discover features shared by all predictive models, but lose these features as the LM explores more specific clusters. We demonstrate that different aspects of linguistic structure are learned at different rates within a single recurrent layer, acquiring POS tags early but continuing to learn global topic information later in training.

\section{Methods}

Our experiments require a LM, tagging models, and a method to inspect the models: SVCCA.

\subsection{Language model}

We model the probability distribution over a sequence of tokens $x_1 \dots x_{|x|}$ with a conventional two-layer LSTM LM. The pipeline from input $x_t$ at time step $t$ to a distribution over $x_{t+1}$ is described in Formulae~(\ref{lm_in})--(\ref{lm_out}). At time step $t$, input word $x_t$ is embedded as (\ref{lm_in}) $h_t^{e}$, which is input to a two-layer LSTM, producing outputs (\ref{lm_lstm1}) $h_{t}^{1}$ and (\ref{lm_lstm2}) $h_{t}^{2}$ at these layers, along with cell states $c_{t}^{1}$ and $c_{t}^{2}$. A softmax layer converts $h_{t}^{2}$ to a distribution from which (\ref{lm_out}) $\hat{x}_{t+1}$ is sampled.
\begin{eqnarray}
h_t^{e} &=& \textrm{embedding}(x_{t})  \label{lm_in} \\
h_t^{1}, c_t^{1} &=& \textrm{LSTM}_{1}(h_t^{\emph{e}}, h_{t-1}^{1}, c_{t-1}^{1}) \label{lm_lstm1} \\
h_t^{2}, c_t^{2} &=& \textrm{LSTM}_2(h_t^{1}, h_{t-1}^{2}, c_{t-1}^{2})\label{lm_lstm2} \\
\hat{x}_{t+1} &\sim& \textrm{softmax}(h_{t}^{2}) \label{lm_out}
\end{eqnarray}
Each function can be thought of as a \emph{representation} or \emph{embedding} of its discrete input; hence $h_t^{e}$ is a representation of $x_t$, and---due to the recursion in (\ref{lm_lstm1})---$h_t^{1}$ is a representation of $x_1 \dots x_t$.

\subsection{Tagging models}

To inspect our language model for learned linguistic categories, we will use a collection of tagging models, designed to mimic the behavior of our language model but predicting the next \emph{tag} rather than the next word. Given $x_1 \dots x_{|x|}$, we model a corresponding sequence of tags $y_1 \dots y_{|x|}$ using a one-layer LSTM. (Our limited labeled data made this more accurate on topic tagging than another two-layer LSTM, so this architecture does not directly parallel the LM.)
\begin{eqnarray}
h_t^e{}' &=& \textrm{embedding}'(x_{t}) \\
h_t^1{}', c_t^1{}' &=& \textrm{LSTM}'(h_t^e{}', h_{t-1}^1{}', c_{t-1}^1{}') \\
\hat{y}_{t+1} &\sim & \textrm{softmax}'(h_t^1{}')
\label{eq:tag_predictor_output}
\end{eqnarray}
We will also discuss \emph{input taggers}, which share this architecture but instead sample $y_t$, the tag of the most recently observed word. 

\subsection{SVCCA} \label{sec:svcca}


SVCCA is a general method to compare the correlation of two vector representations. Let $d_A$ and $d_B$ be their dimensions. For $N$ data points we have two distinct views, given by matrices $A \in \mathcal{R}^{N \times d_A}$ and $B \in \mathcal{R}^{N \times d_B}$. We project these views onto a shared subspace in two steps:

\begin{enumerate}
\item Use Singular Value Decomposition (SVD) to reduce matrices $A$ and $B$ to lower dimensional matrices $A'$ and $B'$, respectively. This is necessary because many dimensions in the representations are noisy, and in fact cancel each other out \citep{frankle_lottery_2018}. SVD removes dimensions that were likely to be less important in the original representations from $A$ and $B$, and in keeping with \citet{raghu_svcca:_2017}, we retain enough dimensions to keep 99\% of the variance in the data.
\item Use Canonical Correlation Analysis (CCA) to project $A'$ and $B'$ onto a shared subspace, maximizing the correlation of the projections. Formally, CCA identifies vectors $w,v$ to maximize $\rho = \frac{<w^\top A', v^\top B'>}{\| 
w^\top A' \|\| v^\top B' \|}$. We treat these $w,v$  as new basis vectors, computing the top $d_C$ (a  hyperparameter) such basis vectors  to form projection matrices $W \in \mathcal{R}^{d_C \times d_{A'}}, V \in \mathcal{R}^{d_C \times d_{A'}} $. The resulting projections $WA'$ and $VB'$  map onto a shared subspace where the representations of each datapoint from $A'$ and $B'$ are maximally correlated.
\end{enumerate}

\begin{figure*}\begin{center}
\begin{tikzpicture}[
node distance=6mm,
function node/.style={fill,minimum size=1mm},
vector/.style={matrix of nodes,draw,nodes={draw,circle,inner sep=0mm},inner sep=0.5mm,column sep=0.5mm,ampersand replacement=\&},
svd/.style={double,blue}
]

\let\fullvector\empty
\foreach \i in {0,...,7} {%
  \expandafter\gappto\expandafter\fullvector\expandafter{~ \&}%
}

\let\reducedvector\empty
\foreach \i in {0,...,4} {%
  \expandafter\gappto\expandafter\reducedvector\expandafter{~ \&}%
}

\node (x_t) {$x_t$};
\node[function node,above of=x_t,label=right:{embedding}] (lm_in) {};
\matrix (h_e) [vector,above of=lm_in,label=left:{$h_t^e$}]{\fullvector \\};
\node[function node,above of=h_e,label=right:{\textsc{Lstm}}] (lm_lstm1) {};
\matrix (h_1) [vector,above of=lm_lstm1,label=left:{$h_t^1$}]{\fullvector \\};
\node[function node,above of=h_1,label=right:{\textsc{Lstm}}] (lm_lstm2) {};
\matrix (h_2) [vector,above of=lm_lstm2,label=left:{$h_t^2$}]{\fullvector \\};
\node[function node,above of=h_2,label=right:{softmax}] (lm_softmax) {};
\node[above of=lm_softmax] (x_tplus1) {$x_{t+1}$};
\node[above of=x_tplus1] (lm_label) {Language model};

\coordinate[left of=lm_lstm1,node distance=15mm] (lm_lstm1_in);
\coordinate[left of=lm_lstm2,node distance=15mm] (lm_lstm2_in);

\path[->]
(x_t) edge (lm_in)
(lm_in) edge (h_e)
(h_e) edge (lm_lstm1)
(lm_lstm1) edge (h_1)
(h_1) edge( lm_lstm2)
(lm_lstm2) edge (h_2)
(h_2) edge (lm_softmax)
(lm_lstm1_in) edge (lm_lstm1)
(lm_lstm2_in) edge (lm_lstm2)
;

\node[right of=lm_in,node distance=11cm] (x_t') {$x_t$};
\node[function node,above of=x_t',label=right:{embedding}] (tagger_in) {};
\matrix (h_e') [vector,above of=tagger_in,label=right:{$h_t^e{}'$}]{\fullvector \\};
\node[function node,above of=h_e',label=right:{\textsc{Lstm}}] (tagger_lstm) {};
\matrix (h_1') [vector,above of=tagger_lstm,label=right:{$h_t^1{}'$}]{\fullvector \\};
\node[function node,above of=h_1',label=right:{softmax}] (lm_softmax) {};
\node[above of=lm_softmax] (y_tplus1) {$y_{t+1}$};
\node[above of=y_tplus1] (tagger_label) {Tag Predictor};

\coordinate[left of=tagger_lstm,node distance=10mm] (tagger_lstm_in);

\path[->]
(x_t') edge (tagger_in)
(tagger_in) edge (h_e')
(h_e') edge (tagger_lstm)
(tagger_lstm) edge (h_1')
(h_1') edge (lm_softmax)
(lm_softmax) edge (y_tplus1)
(tagger_lstm_in) edge (tagger_lstm)
;

\coordinate[right of=h_2,xshift=5mm] (lm_svd_start);
\coordinate[right of=lm_svd_start,node distance=2cm] (lm_svd_end);
\matrix (svd_h_2) [vector,right of=lm_svd_end,label=above:{SVD($h^2$)}]{\reducedvector \\};

\coordinate[left of=h_1',xshift=-5mm] (tagger_svd_start);
\coordinate[left of=tagger_svd_start,node distance=2cm] (tagger_svd_end);
\matrix (svd_h_1') [vector,left of=tagger_svd_end,label=above:{SVD($h^1{}'$)}]{\reducedvector \\};

\path[double,->]
(lm_svd_start) edge[svd] node[above] {SVD} (lm_svd_end)
(tagger_svd_start) edge[svd] node[above] {SVD} (tagger_svd_end)
;

\coordinate[right of=lm_in,node distance=3.7cm] (cca_sw);
\coordinate[above right of=cca_sw,node distance=2cm] (cca_nw);
\coordinate[right of=cca_sw,node distance=2cm] (cca_se);
\coordinate[right of=cca_nw,node distance=2cm] (cca_ne);

\coordinate[left of=cca_se,xshift=5mm,label=below:{$\displaystyle\max_{W,V} \rho(W\cdot\textrm{SVD}(h^2),V\cdot\textrm{SVD}(h^1{}'))$}] (cca_label);

\coordinate[below of=svd_h_2,node distance=3mm] (cca_h_2_st);
\coordinate[below of=cca_nw,xshift=2mm] (cca_h_2_en);
\node[below of=cca_h_2_en,node distance=1mm,circle,fill,red,inner sep=0.4mm] (cca_h_2) {};

\coordinate[below of=svd_h_1',node distance=3mm] (cca_h_1_st');
\coordinate[right of=cca_h_2_en,xshift=-3mm,yshift=2mm] (cca_h_1_en');
\node[below of=cca_h_1_en',node distance=1mm,circle,fill,red,inner sep=0.4mm] (cca_h_1') {};

\path
(cca_sw) edge[thick,red] (cca_nw)
(cca_nw) edge[thick,red] (cca_ne)
(cca_ne) edge[thick,red] (cca_se)
(cca_se) edge[thick,red] (cca_sw)
(cca_h_2_st) edge[->,double,red] (cca_h_2_en)
(cca_h_1_st') edge[->,double,red] (cca_h_1_en')
;

\end{tikzpicture}
\caption{SVCCA used to compare the layer $h^2$ of a language model and layer $h^1{}'$ of a tagger.}
\label{fig:procedure}\end{center}
\end{figure*}
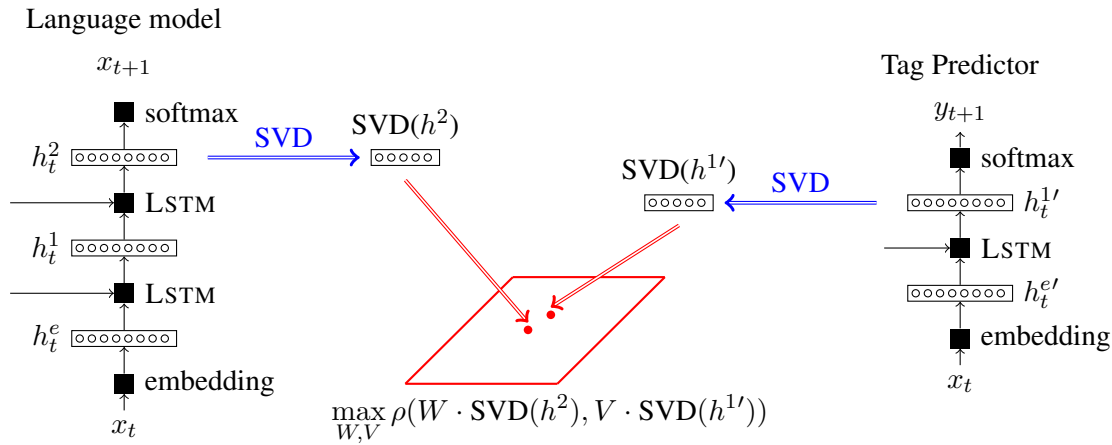

Intuitively, the correlation $\rho$ will be high if both representations encode the same information, and low if they encode unrelated information. Figure~\ref{fig:procedure} illustrates how we use SVCCA to compare representation $h_t^2$ of our language model with the recurrent representation of a tagger, $h_t^1{}'$. In practice, we run over all time steps in a test corpus, rather than a single time step as illustrated.

\section{Experimental Setup}

\begin{table*}[ht]
\centering
\begin{tabular}{c | llllllll}
Tag & These & cats & live & in & that & house & . \\
\hline
UDP POS & DET & NOUN & VERB & ADP & DET & NOUN & SYM \\
PTB POS & DT & NNS & VBP & IN & DT & NN & . \\
SEM (coarse) & DEM & ENT & EVE & ATT & DEM & ENT & LOG \\
SEM (fine) & PRX & CON & ENS & REL & DST & CON & NIL \\
topic & pets & pets & pets & pets & pets & pets & pets \\
\end{tabular}
\caption{An annotated example sentence from the article \textit{pets}, based on an example from \citet{bjerva_semantic_2016}.}
\label{fig:example}
\end{table*}

We trained our LM on a corpus of tokenized, lowercased English Wikipedia (70/10/20 train/dev/test split). To reduce the number of unique words in the corpus, we excluded any sentence with a word type appearing fewer than 100 times. Words appearing fewer than 100 times in the resulting training set are replaced with an unknown token. The resulting training set has over 227 million tokens of 20K types.

\begin{figure}
\includegraphics[width=0.48\textwidth]{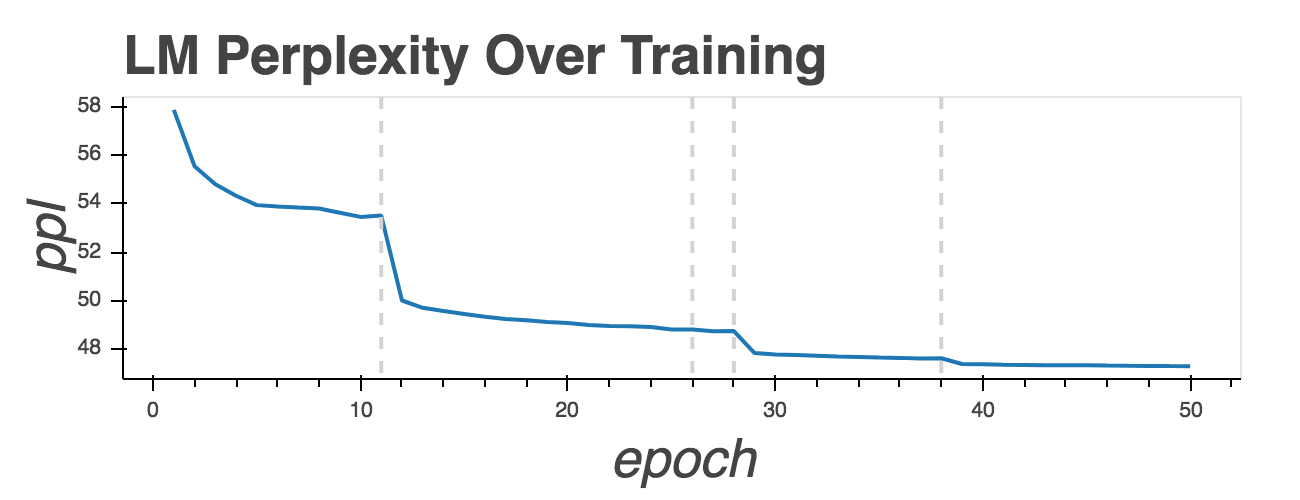}
\caption{Test performance of the LM. Vertical dotted lines indicate when the optimizer rescales the step size.}
\label{fig:loss}
\end{figure}

We train for 50 epochs to maximize cross-entropy, using a batch size of 40, dropout ratio of 0.2, and sequence length of 35. The optimizer is standard SGD with clipped gradients at 0.25, with the learning rate quartered when validation loss increases. The result of training is shown in Figure~\ref{fig:loss}, which illustrates the dips in loss when learning rate changes. All experiments on the LM throughout training are conducted by running the model at the end of each epoch in inference mode over the test corpus.

\subsection{Taggers}

To understand the representations learned by our LM, we compare them with the internal representations of tagging models, using SVCCA. Where possible, we use coarse-grained and fine-grained tagsets to account for effects from the size of the tagset. Table~\ref{fig:example} illustrates our tagsets.

\paragraph{POS tagging}
For syntactic categories, we use POS tags, as in \citet{belinkov_what_2017}. As a coarse-grained tagset, we use silver Universal Dependency Parse (UDP) POS tags automatically added to our Wikipedia corpus with spacy.\footnote{https://spacy.io/} We also use a corpus of fine-grained human annotated Penn Treebank POS tags from the Groningen Meaning Bank \citep[GMB;][]{bos2017groningen}.

\paragraph{Semantic tagging} We follow \citet{belinkov_evaluating_2018} in representing word-level semantic information with silver SEM tags \citep{bjerva_semantic_2016}. SEM tags disambiguate POS tags in ways that are relevant to multilingual settings. For example, the comma is not assigned a single tag as punctuation, but has distinct tags according to its function: conjunction, disjunction, or apposition. The 66 fine-grained SEM tag classes fall under 13 coarse-grained tags, and an `unknown' tag.

\paragraph{Global topic} For topic, we classify each word of the  sequence by its source Wikipedia article; for example, every word in the wikipedia article on Trains is labeled ``Trains''. This task assesses whether the network encodes the global topic of the sentence. 

UDP silver POS and topic information use the same corpus, taken from the 100 longest articles in Wikipedia randomly partitioned in a 70/10/20 train/dev/test split.   Each token is tagged with POS and with the ID of the source article. The corpus is taken from the LM training data, which may increase the similarity between the tag model and LM. Because both tag predictors are trained and tested on the same domain as the LM, they can be easily compared in terms of their similarity to the LM representation. Though the SEM corpus and the PTB corpus are different domains from the Wikipedia training data, we compare their activations on the same 191K-token 100-article test corpus. 

Table~\ref{fig:corpus_stats} describes the training and validation corpus statistics for each tagging task. Note that topic and UDP POS both apply to the same en-wikipedia corpus, but PTB POS and SEM use two different unaligned sets from the GMB corpus.

\begin{table*}
\centering\small
\begin{tabular}{l||l|r|rr|rr|rr|rr}
 &  & \multicolumn{1}{|l|}{number}  & \multicolumn{2}{|c|}{token count}  & \multicolumn{2}{|c|}{label $t+1$} & \multicolumn{2}{|c|}{label $t$} & \multicolumn{2}{|c}{randomized}\\
tag & corpus & of classes & train & dev  & acc & ppl  & acc & ppl & acc & ppl \\
\hline
UDP POS & wiki & 17 & 665K & 97K &  50 & 4.3    & 93 & 1.2 & 21 & 8.9\\
PTB POS & GMB & 36 & 943K & 136K & 51 & 4.7 & 95 & 1.18 & 14 & 18.0 \\
SEM (coarse) & GMB & 14 & 937K & 132K &  55 & 3.5 & 91 & 1.3 & 22 & 9.0 \\
SEM (fine) & GMB & 67 & 937K & 132K &  50 & 5.6 & 88 & 1.45 & 17 & 21.5   \\
topic & wiki & 100 & 665K & 97K  & 36 & 19.1 & 37 & 16.3 & 5 & 81.6\\
\end{tabular}
\caption{Tag predictor and tagger statistics. Accuracy and perplexity on $t+1$ are from the target tag predictor, on $t$ are from the input tagger. Metrics obtained when training on randomly shuffled labels are provided as a low baseline. Accuracy is on the test set from the training domain (GMB or Wikipedia).}
\label{fig:corpus_stats}
\end{table*}

\section{Experiments, Results, and Analysis}

\begin{figure}
\includegraphics[width=0.5\textwidth]{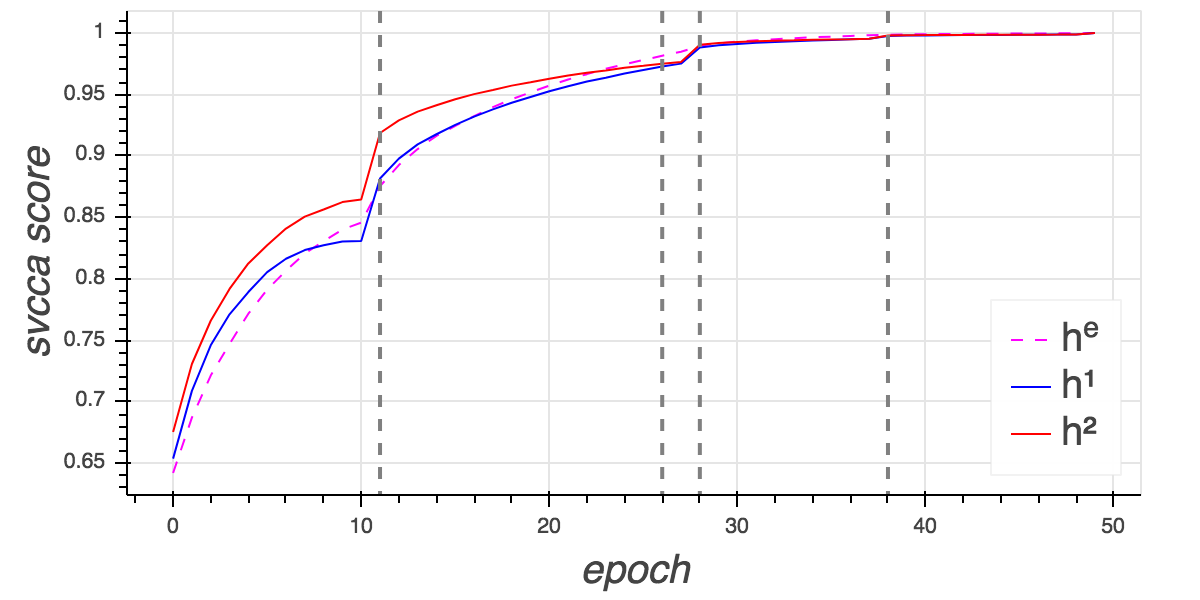}
\caption{SVCCA score between representations at each epoch and from the final trained LM.}
\label{fig:final_convergence}
\end{figure}

A benefit of SVCCA is its flexibility: it can compute the correlation of a hidden representation to any other vector. \citet{raghu_svcca:_2017} used it to understand learning dynamics by comparing a learned representation to snapshots of the same representation at different epochs during training. We use a similar experiment to establish the basic learning dynamics of our model. In our shallow 2-level model, activations at $h^1$ converge slightly after $h^2$ (Figure~\ref{fig:final_convergence}). This differs from the results of \citet{raghu_svcca:_2017}, who found that a 5-layer stacked LSTM LM exhibits faster convergence at lower layers, but this difference may be attributed to our much larger training data, which gives our model sufficient training data at early epochs. 

\begin{figure}
\includegraphics[width=0.5\textwidth]{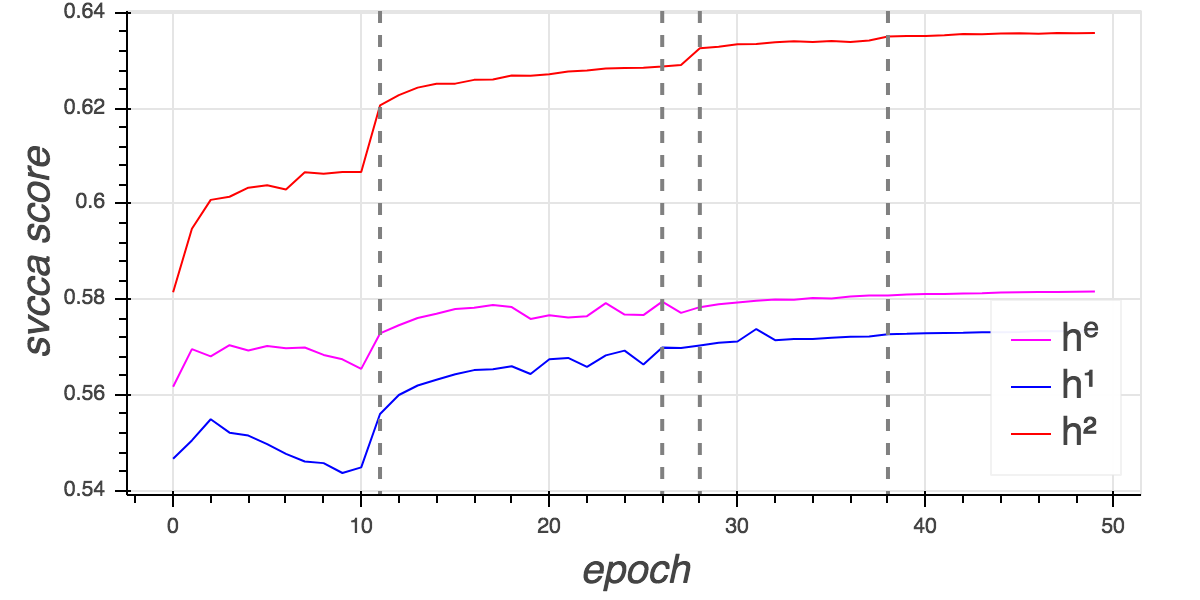}
\caption{SVCCA score between the LM at each epoch and a LM with different initialization.}
\label{fig:random_convergence}
\end{figure}

\paragraph{Empirical upper bounds.} Our main experiments will test the rate at which different linguistic categories are learned by different layers, but to interpret the results, we need to understand the behaviour of SVCCA for these models. In theory, SVCCA scores can vary from 0 for no correlation to 1 for perfect correlation. But in practice, these extreme cases will not occur. To establish an \emph{empirical} upper bound on correlation, we compared the similarity at each epoch of training to the frozen final state of a LM with identical architecture but different initialization, trained on the same data (Figure~\ref{fig:random_convergence}).\footnote{This experiment is similar to the comparisons of randomly initialized models by \citet{morcos_insights_2018}.} The correlations increase over time as expected, but to a maximum near 0.64; we don't expect correlations between our LM and other models to exceed this value. We explore corresponding lower bounds in our main experiments below.

\begin{figure}
\includegraphics[width=0.5\textwidth]{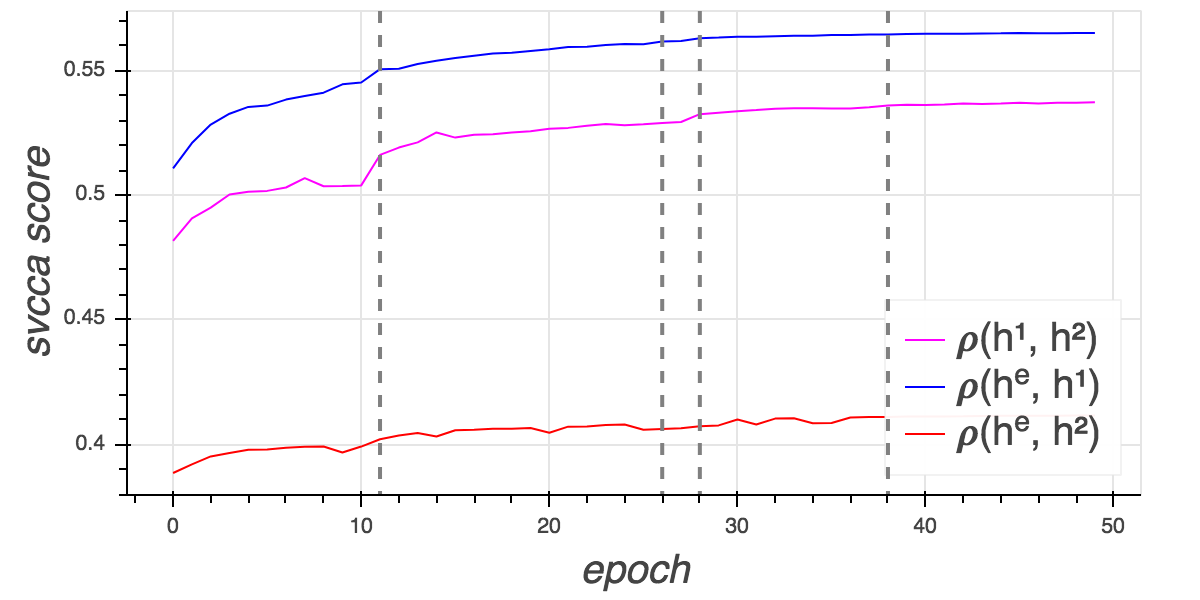}
\caption{SVCCA score between different layers of the LM at each epoch. For example, $\rho({h^2, h^1})$ compares the activations $h^2$ with the activations $h^1$.}
\label{fig:between_layers}
\end{figure}

\paragraph{Correlations between different layers.} Next we examine the correlation between different layers of the same model over time (Figure~\ref{fig:between_layers}). We observe that, while over time correlation increases, in general closer layers are more similar, and they are less correlated than they are with the same layer of a differently initialized model. This supports the idea that we should compare  recurrent layers with recurrent layers because their representations play similar roles within their respective architectures.

\paragraph{SVCCA vs. Diagnostic classifiers}

A popular method to analyze learned representations is to use a \emph{diagnostic classifier} \citep{belinkov_what_2017}, a separate model that is trained to predict a linguistic category of interest, $y_t$, from an arbitrary hidden layer $h_t$. Diagnostic classifiers are widely used \citep{belinkov_evaluating_2018,giulianelli_under_2018}. But if a diagnostic classifier is trained on enough examples, then random embeddings as input representations often outperform any pretrained intermediate representation \citep{wieting2019no,zhang_language_2018}. This suggests that diagnostic classifiers may work simply by memorizing the association between an embedding and the most frequent output category associated with that embedding; since for many words their category is (empirically) unambiguous, this may give an inflated view of just how much a model ``understands'' about that category. 

Our use of SVCCA below will differ from the use of diagnostic classifiers in an important way.  Diagnostic classifiers use the intermediate representations of the LM as inputs to a tagger. A representation is claimed to encode, for example, POS if the classifier accurately predicts it---in other words, whether it can \emph{decode} it from the representation. We will instead evaluate the \textit{similarity} between the representations in an LM and in an independently-trained tagger. The intuition behind this is that, if the representation of our LM encodes a particular category, then it must be similar to the representation of model that is specifically trained to predict that category. A benefit of the approach is that similarity can be evaluated on \emph{any} dataset, not only one that has been labeled with the linguistic categories of interest.

Another distinction from the typical use of diagnostic classifiers  is that probes are  usually used to decode tag information about the context or most recent \textit{input} from the hidden state at the current step. Because the hidden representation at time $t$ is meant to encode predictive information about the target word at time $t+1$, we treat it as encoding a prediction about the tag of the \textit{target} word.

\begin{figure}
\includegraphics[width=0.5\textwidth]{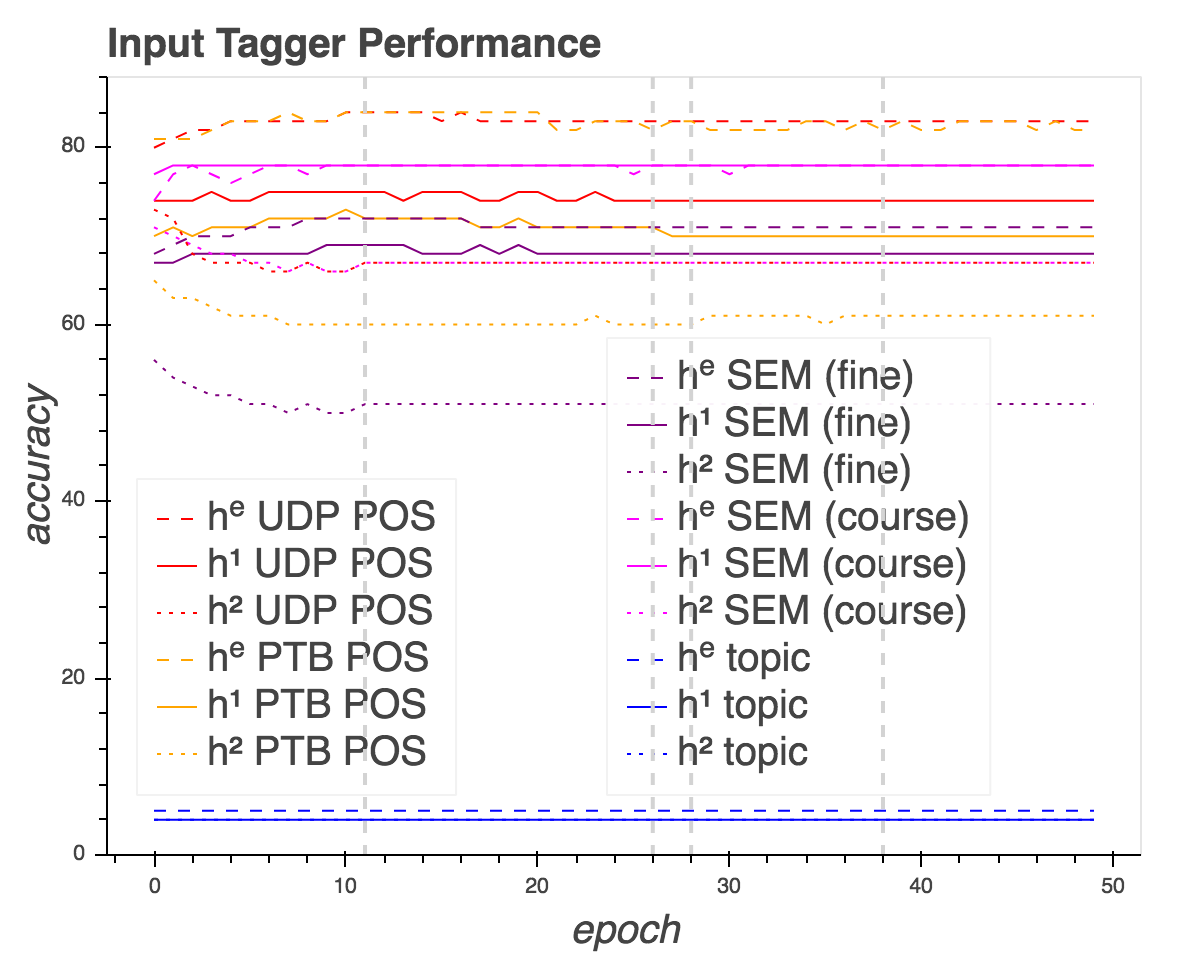}
\caption{Learning dynamics interpreted with diagnostic classifiers labeling input word tag $y_{t}$.}
\label{fig:belinkov_input}
\end{figure}

\begin{figure}
\includegraphics[width=0.5\textwidth]{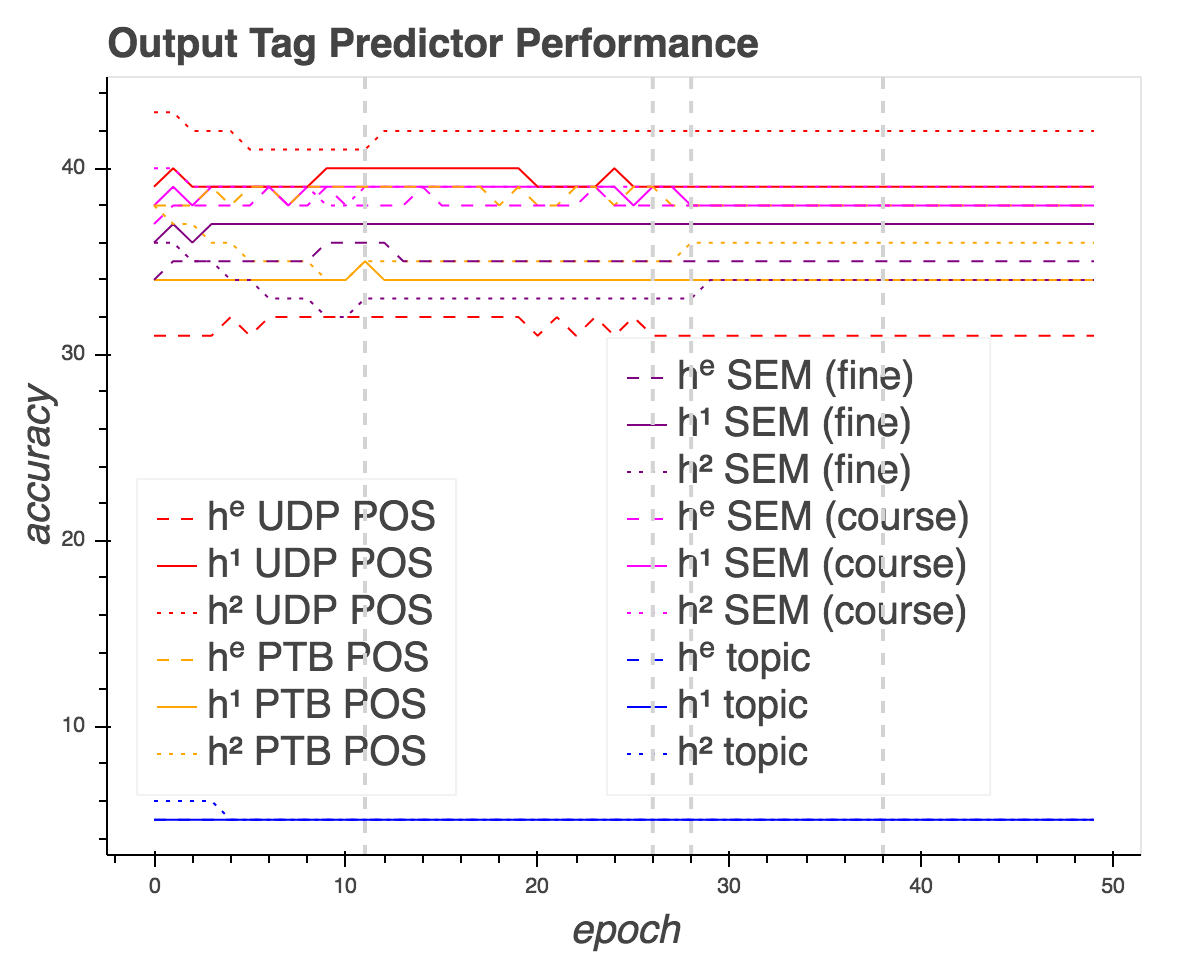}
\caption{Learning dynamics interpreted with diagnostic classifiers labeling target word tag $y_{t+1}$.}
\label{fig:belinkov}
\end{figure}

To understand the empirical strengths and weaknesses of these approaches, we compare the use of SVCCA and diagnostic classifiers in understanding learning dynamics. In other words, we ask: is our first conceptual shift (to SVCCA) necessary? To test this, we use the same model as \citet{belinkov_what_2017}, which classifies an arbitrary representation using a ReLU followed by a softmax layer. To be consistent with \citet{belinkov_what_2017}, we use $y_t$ as their target label. We repeat their method in this manner (Figure~\ref{fig:belinkov_input}) as well as applying our second modification, in which we instead target the label $y_{t+1}$ (Figure~\ref{fig:belinkov}).

We found the correlations to be relatively stable over the course of training. This is at odds with the results in Figures~\ref{fig:loss} and \ref{fig:final_convergence}, which suggest that representations change substantially during training in ways that materially affect the accuracy of the LM. This suggests that diagnostic classifiers are not illustrating improvements in word representations throughout training, and we conclude that they are ineffective for understanding learning dynamics. Our remaining experiments use only SVCCA.

\subsection{SVCCA on Output Tag Prediction}

\begin{figure}
\includegraphics[width=0.49\textwidth]{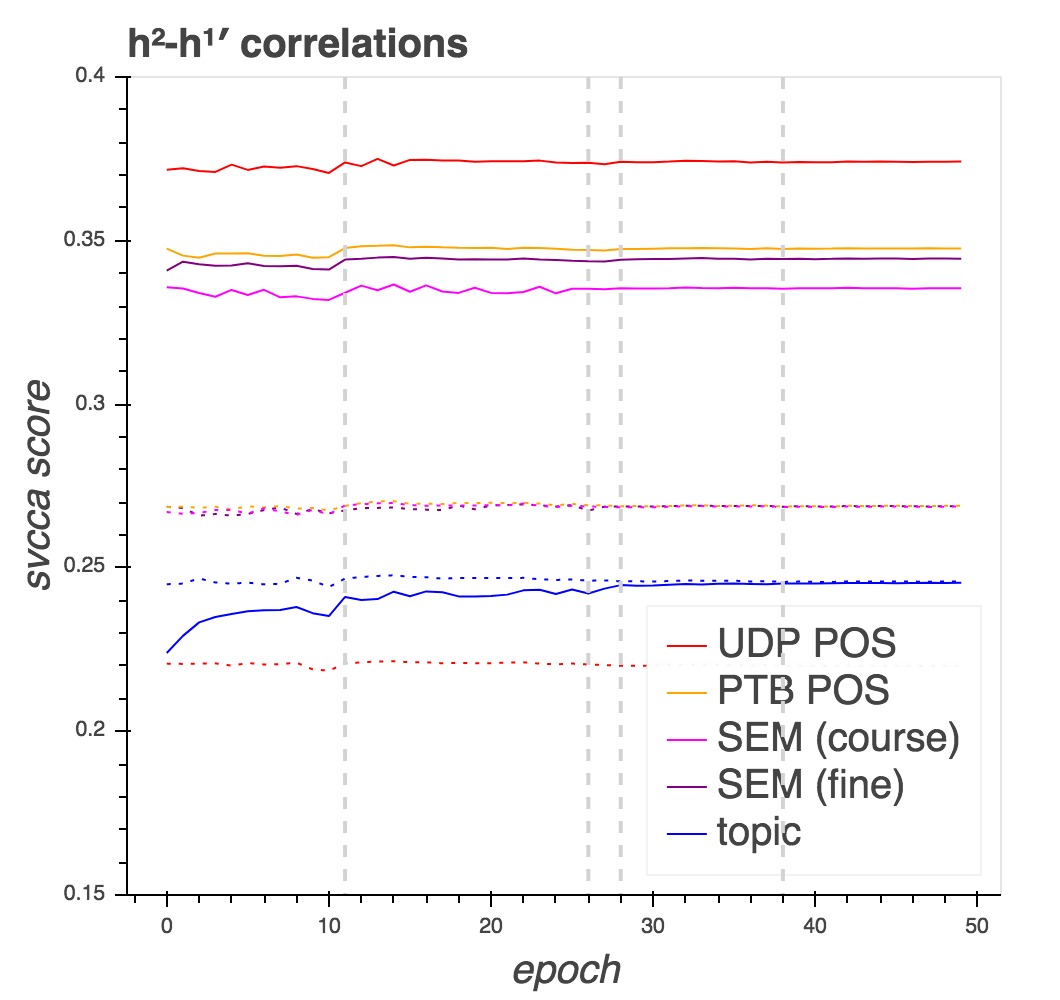}
\includegraphics[width=0.49\textwidth]{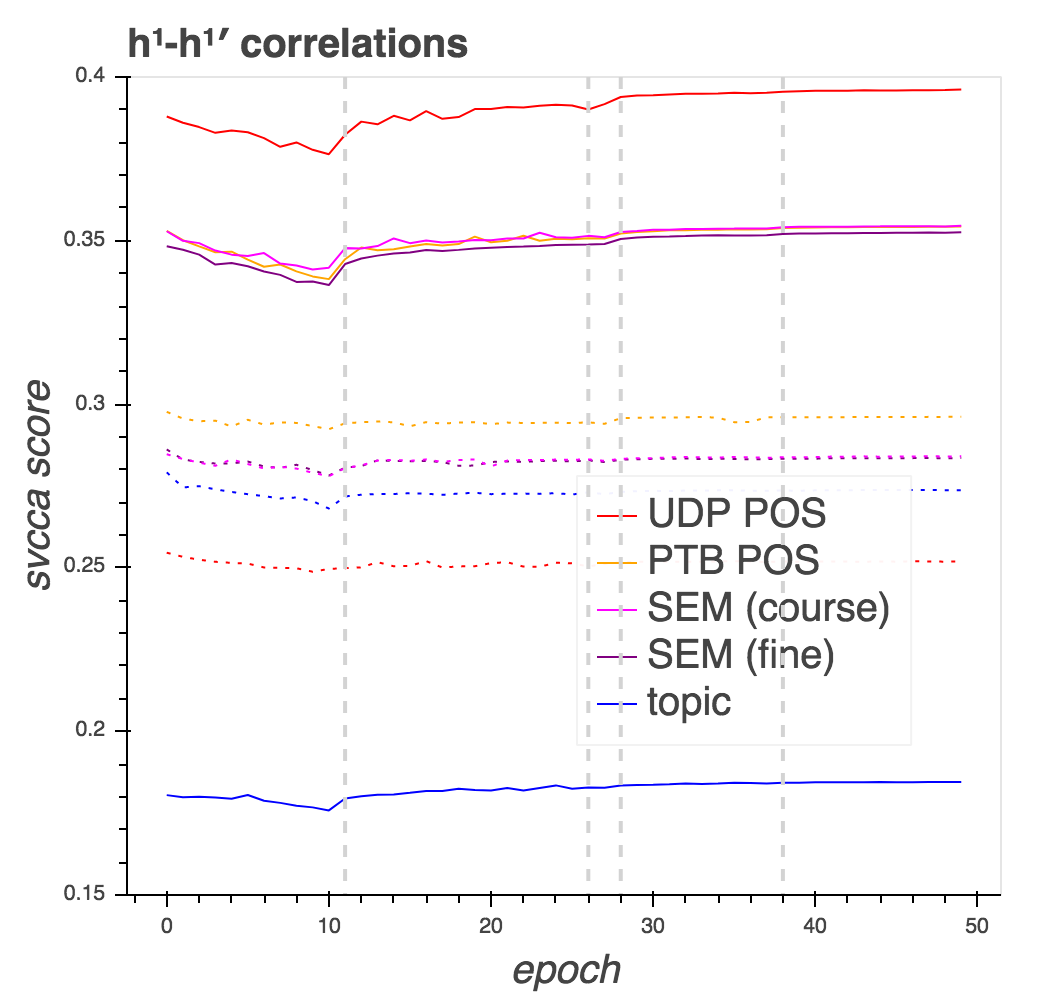}
\includegraphics[width=0.49\textwidth]{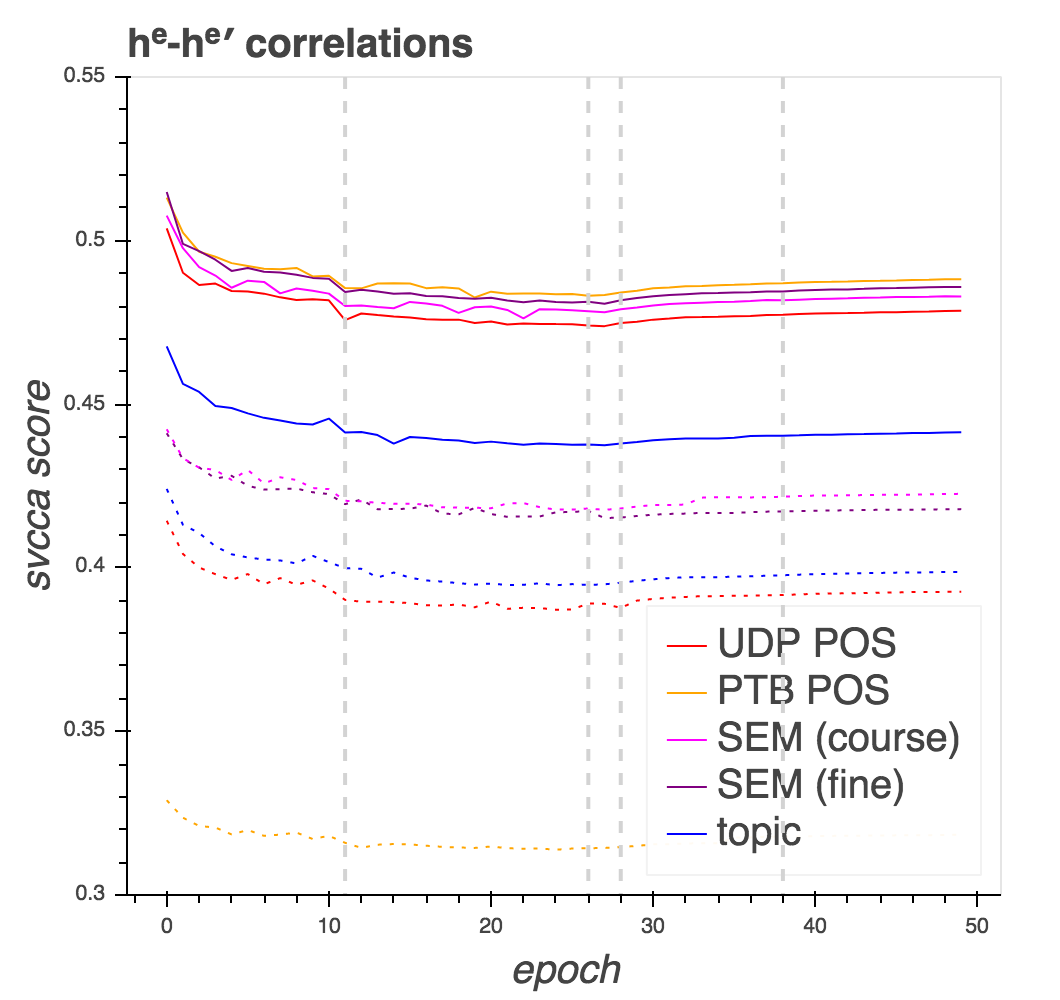}
\caption{SVCCA correlation scores between the LM predicting $x_{t+1}$ and the tag model predicting $y_{t+1}$.  At the end of each epoch, we compare the current LM with the final tag model. Dotted lines use shuffled tags. Gray vertical lines mark when the step size is rescaled.}
\label{fig:svcca_encoder_rnn_1}
\end{figure}

We applied SVCCA to each layer of our LM with the corresponding layer of each tag predictor in order to find the correlation between the LM representation and the tag model representation at each level (Figure~\ref{fig:svcca_encoder_rnn_1}).
To establish empirical lower bounds on correlation, we also trained our taggers on the same data with randomly shuffled labels, as in \citet{zhang2016understanding}. These latter experiments, denoted by the dotted lines of Figure~\ref{fig:svcca_encoder_rnn_1}, show how much of the similarity between models is caused by their ability to memorize arbitrary associations. Note that the resulting scores are  nonzero, likely because the linguistic structure of the input shapes representations even when the output is random, due to the memorization phase of training \citep{shwartz-ziv_opening_2017}.

The strongest similarity at recurrent layers belongs to the most local property, the UDP POS tag. Both coarse- and fine-grained semantic tags, which rely on longer range dependencies, fall below UDP POS consistently. Topic, which is global to an entire document, is the least captured and the slowest to stabilize. Indeed, correlation with true topic falls consistently \emph{below} the score for a model trained on randomized topic tags, implying that early in training the model has  removed the context necessary to identify topic (below even the inadequate contextual information memorized by a model with random labels), which depends on the general vocabulary in a sentence rather than a local sequence. Over time correlation rises, possibly because the model permits more long-distance context to be encoded. \citet{khandelwal_sharp_2018} found that LSTMs remember content words like nouns for more time steps than they remember function words like prepositions and articles. We hypothesize that the LM's slower stabilization on topic is related to this phenomenon, since it must depend on content words, and its ability to remember them increases throughout training.

The encoder layer exhibits very different patterns. Because the representation produced by the encoder layer is local to the word, the nuances that determine how a word is tagged in context cannot be learned. The encoder layers are all highly similar to each other, which suggests that the unigram representations produced by the encoder are less dependent on the particular end task of the neural network. Similarity between the encoders declines over time as they become more specialized towards the language modeling task. This decline points to some simple patterns which are learned for all language tasks, but which are gradually replaced by representations more useful for language modeling. This process may even be considered a naturally occurring analog to the common practice of initializing the encoder layer as word embeddings pretrained an unrelated task  such as skipgram or CBOW \citep{mikolov_distributed_2013}. It seems that the `easy' word properties, which immediately improve performance, are similar regardless of the particular language task. 

At $h^1$, the correlation shows a clear initial decline in similarity for all tasks. This seems to point to an initial representation that relies on simple shared properties, which in the first stage of training is gradually dissolved before the layer begins to converge on a structure shared with each tag predictor. It may also be linked to the information bottleneck learning phases explored by  \citet{shwartz-ziv_opening_2017}. They suggest that neural networks learn by first maximizing the mutual information between the input and internal representation, then minimizing the mutual information between the internal representation and output. The network thus initially learns to effectively represent the input, then compresses this representation, keeping only the elements relevant to the output.\footnote{This memorization–compression learning pattern parallels the memorization–generalization  of the first half of the U-shaped curve exhibited by  human children learning irregular word forms. \citet{kirov2018recurrent} observe similar patterns when artificially modeling inflection.} If the LM begins by maximizing mutual information with input, because the input is identical for the LM and tag models it may lead to these similar initial representations, followed by a decline in similarity as the compression narrows to properties specific to each task.

\subsection{SVCCA on Input Tagging}

\begin{figure}
    \centering
    \includegraphics[width=0.5\textwidth]{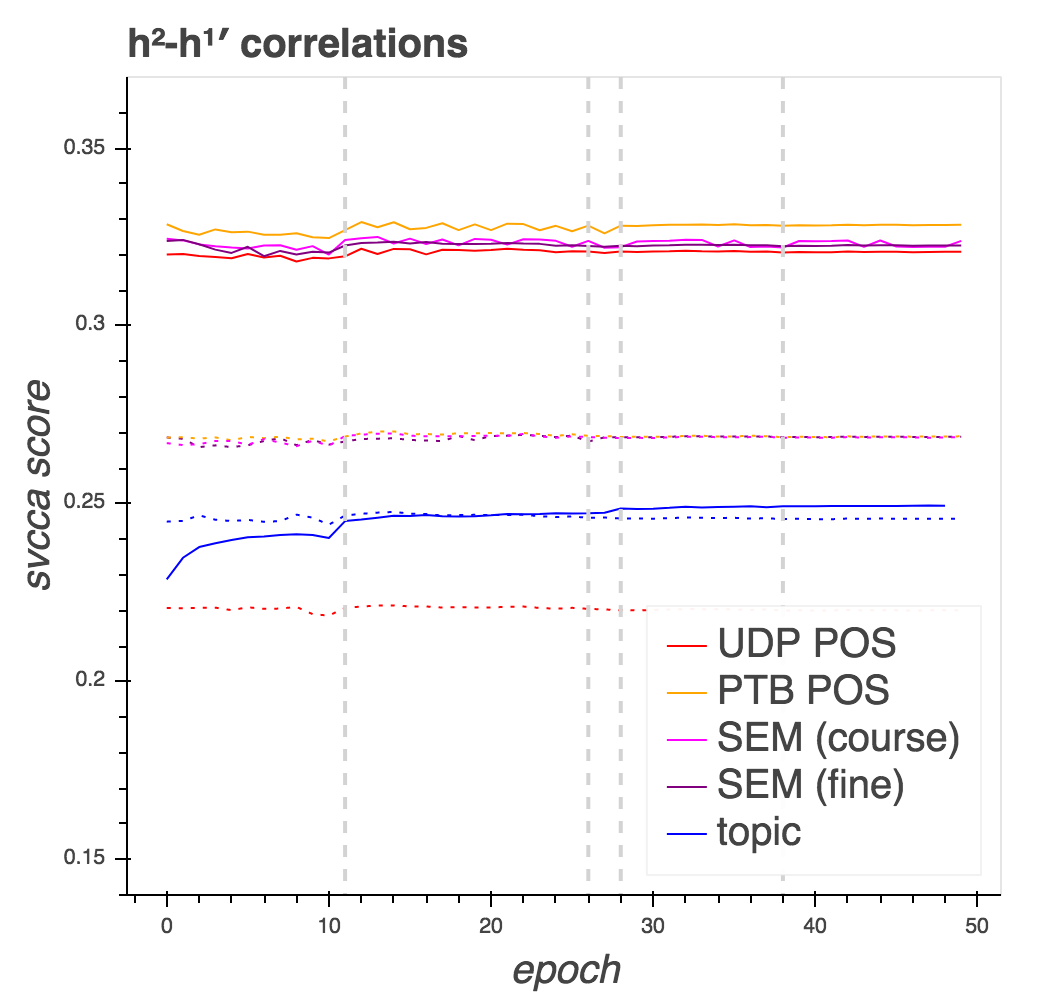}
    \includegraphics[width=0.5\textwidth]{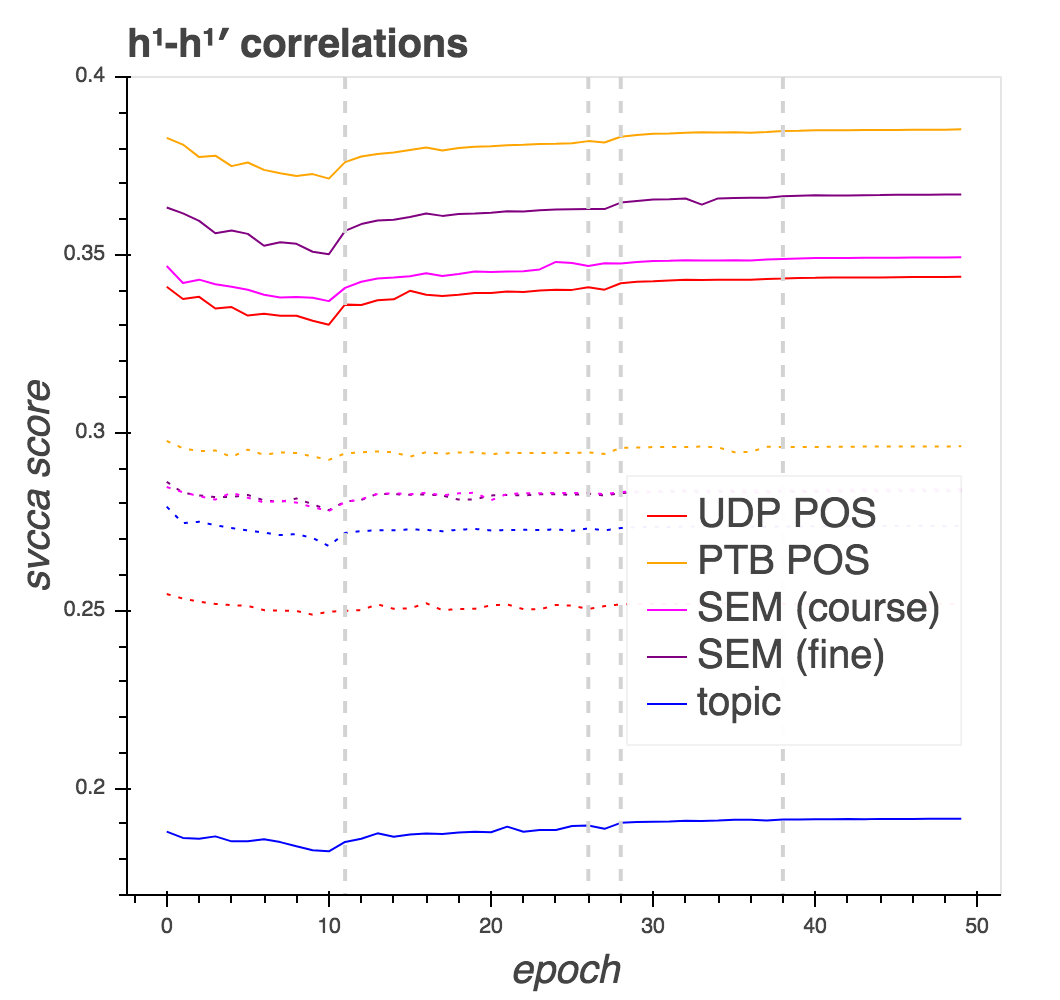}
    \includegraphics[width=0.5\textwidth]{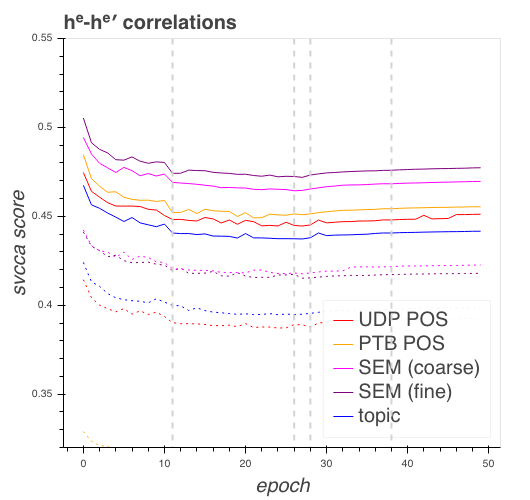}
    \caption{SVCCA correlation scores between LM activations when predicting $x_{t+1}$ and tagger activations when labeling $y_t$. Dotted lines use shuffled tags. Gray vertical lines mark when the step size is rescaled.}
    \label{fig:input_cca}
\end{figure}

Our second conceptual shift is to focus on output tag prediction---asking what a representation encodes about the \emph{next} output word, rather than what it has encoded about words it has already observed in the input. What effect does this have? Since we already studied output tags in the previous set of experiments, here we consider input tags, in the style of most diagnostic classifier analysis (Figure~\ref{fig:input_cca}). The learning dynamics are similar to those for tag prediction, but the UDP POS tagger decreases dramatically in all correlations while the GMB-trained taggers\footnote{PTB POS, SEM (fine), and SEM (coarse)} often increase slightly. While the shapes of the lines are similar, UDP POS no longer consistently dominates the other tasks in recurrent layer correlation. Instead, we find the more granular PTB POS tags lead to the most similar representations. 

\section{Discussion and Conclusions}

We find clear patterns in the encoding of linguistic structure with SVCCA, in contrast to the weaker results from a less responsive diagnostic classifier. Because SVCCA proves so much more sensitive than the diagnostic classifiers currently in use, we believe that future work on measuring the encoding of linguistic structure should use the similarity of individual modules from independently trained tag predictors rather than the performance of tag predictors trained on a particular representation.

This system should also be of interest because it is efficient. To train a diagnostic classifier, we must run a forward pass of the LM for each forward pass of the auxiliary model,  while SVCCA only requires the LM  to run on the test set. A further efficiency gain is  particular  to studying learning dynamics: we train only one tagger and compare it to   different versions of the LM   over training, but  for standard probing, we must  train  a new  version of each layer's tagger at each epoch. Our SVCCA experiments  in Figure~\ref{fig:svcca_encoder_rnn_1} ran in hours, while the diagnostic classifier experiments in Figure~\ref{fig:belinkov}  ran for days.

Our method holds another, more subtle advantage. Our analysis provides an alternative view of what it means  for a model to \textit{encode} some linguistic trait. The literature on analyzing neural networks includes a broad spectrum of interpretations about what it means to encode a property. At one end of the spectrum lies  the purely informational view \citep[e.g., mutual information;][]{noshad_scalable_2018}. Mutual information is a very flexible view, but it  requires us to compute theoretical information content, which in practice can only be estimated. Furthermore, information can be represented without being used, as shown by \citet{vanmassenhove2017investigating}, who found that NMT systems often predicted tense according to a diagnostic classifier but did not produce the correct tense as output. The other end of the spectrum is focused on the structure of the representation space \citep[e.g., the features and the property in question are linearly similar;][]{alishahi2017encoding}. Analyzing structural similarity should remedy  the shortcomings of the informational view, but  most intermediate representations  are not targeted  to extract the property in question through a  linear transformation, and failing to be interpretable through  such simple extraction should not be equated  to a failure to encode  that property. 
 
 Most of the literature on analyzing representations, by  probing with a  more complex architecture, seeks  the flexibility of mutual information  with the concreteness  and tractability of the structural view  --  but instead obscures the strict information view without offering interpretable information about the structure, because the architecture of a  diagnostic classifier affects its performance.  It should not be surprising that  representational quality as measured by such systems is   a poor indicator  of translation quality \citep{cifka2018bleu}. SVCCA, in contrast, is a structural view that  does not directly compare an activation  that targets word prediction with a particular tag, but instead compares that activation with one targeting the prediction of the tag. 

Let us consider a specific common probing method. What do we learn about the LM when a feedforward network cannot extract tag information directly from the embedding layer, but can from a recurrent layer? It may be tempting to conclude that tag information relies heavily on context, but consider some alternative explanations. If the embedding encodes the tag to be interpreted by a recurrent layer, a feedforward network may not be capable of representing the function to extract that tag because it does not have access to a context vector for aiding interpretation of the hidden layer. Perhaps its activation functions cover a different range of outputs. By directly comparing LSTM layers to LSTM layers and embedding layers to embedding layers, we respect the shape of their outputs and the role of each module within the network in our analysis. 

The results of our analysis imply that early in training, representing part of speech is the natural way to get initial high performance. However, as training progresses, it increasingly benefits the model to represent categories with longer-range dependencies, such as topic.

\section{Future Work}

One direction for future work is exploring how generalization interacts with the correlations between LMs and tag predictors. It may be that a faithful encoding of a property like POS tag indicates that the LM is relying more on linguistic structure than on memorizing specific phrases, and therefore is associated with a more general model. 

If these measurements of structure encoding are associated with more general models, we might introduce regularizers or other modifications that explicitly encourage correlation with a tagging task.

\citet{combes_learning_2018} identified the phenomenon of \emph{gradient starvation}, meaning that while frequent and unambiguous features are learned quickly in training, they slow down the learning of rarer features. For example, artificially brightening images according to their class leads to a delay in learning to represent the more complex natural class features. Although it is tempting to claim that semantic structure is learned using  syntactic structure as natural scaffolding, it is possible that the simple predictive power of POS is acting as an attractor and starving semantic features that are rarer and more ambiguous. A possible direction for future work would be to explore which of these explanations is true, possibly by decorrelating particular aspects of linguistic structure from language modeling representations.


The techniques in this paper could be applied to better understand the high performance of a system like ELMo \citep{peters2018deep}. Different layers in such a system are useful for different tasks, and this effect could be understood in terms of the gradual divergence between the layers and their respective convergence to representations geared toward a single task. 

\section*{Acknowledgements}
We thank
Denis Emelin,
Sameer Bansal,
Toms Bergmanis,
Maria Corkery,
Sharon Goldwater,
Sorcha Gilroy,
Aibek Makazhanov,
Yevgen Matusevych,
Kate McCurdy,
Janie Sinclair,
Ida Szubert,
Nikolay Bogoychev,
Clara Vania, and the anonymous reviewers
for helpful discussion and comments on drafts.
We thank Matthew Summers for assistance with visualisations.

\bibliography{emnlp2018}
\bibliographystyle{acl_natbib}

\newpage

\appendix

\newpage

\section{Performance Out Of Domain}
\label{sec:domain}

\begin{figure}[hb]
\includegraphics[width=0.5\textwidth]{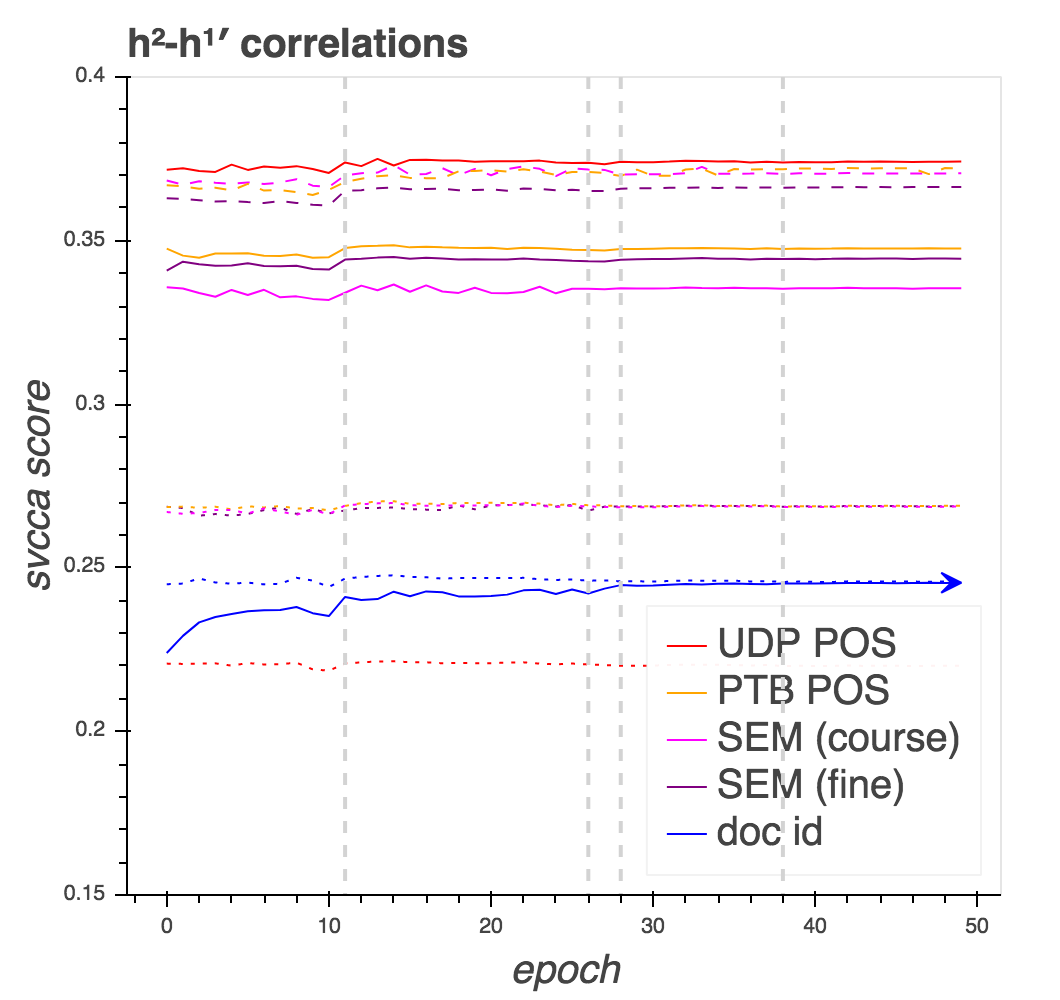}
\includegraphics[width=0.5\textwidth]{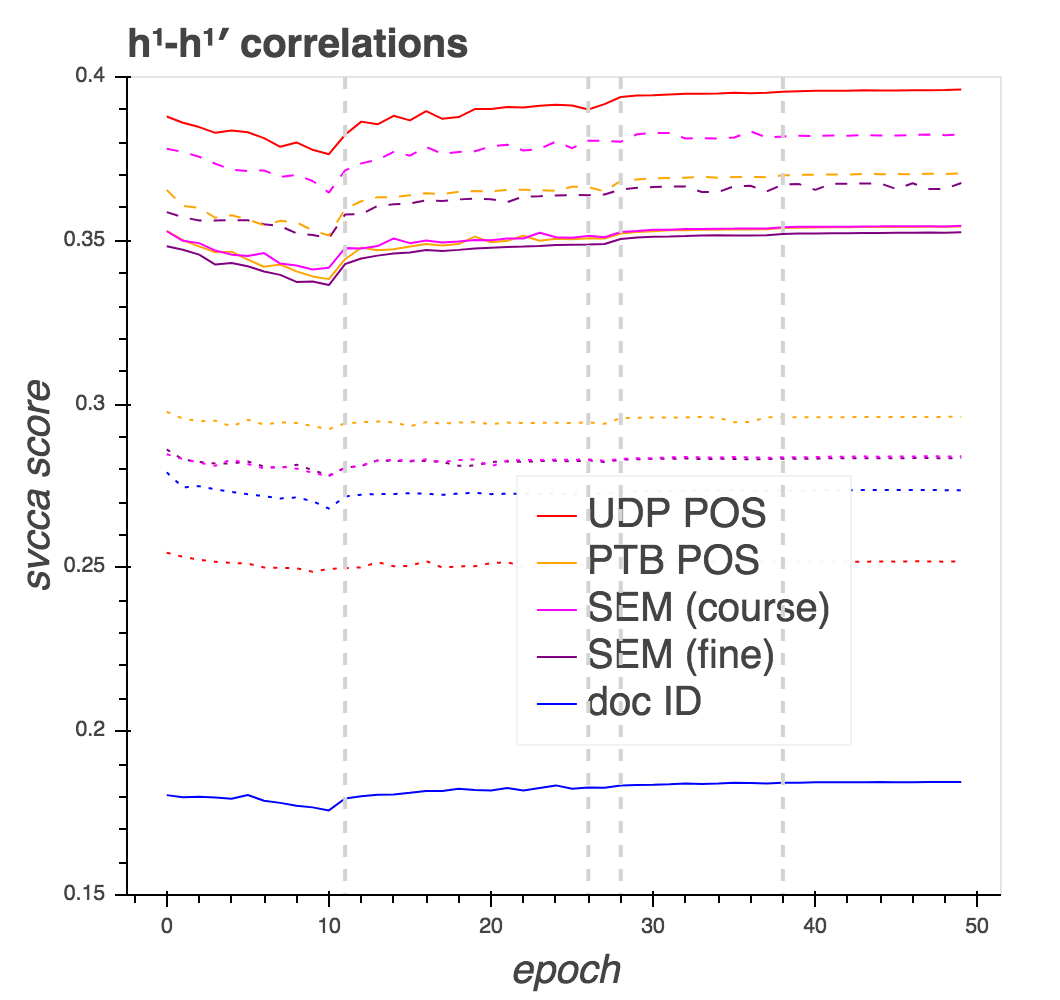}
\includegraphics[width=0.5\textwidth]{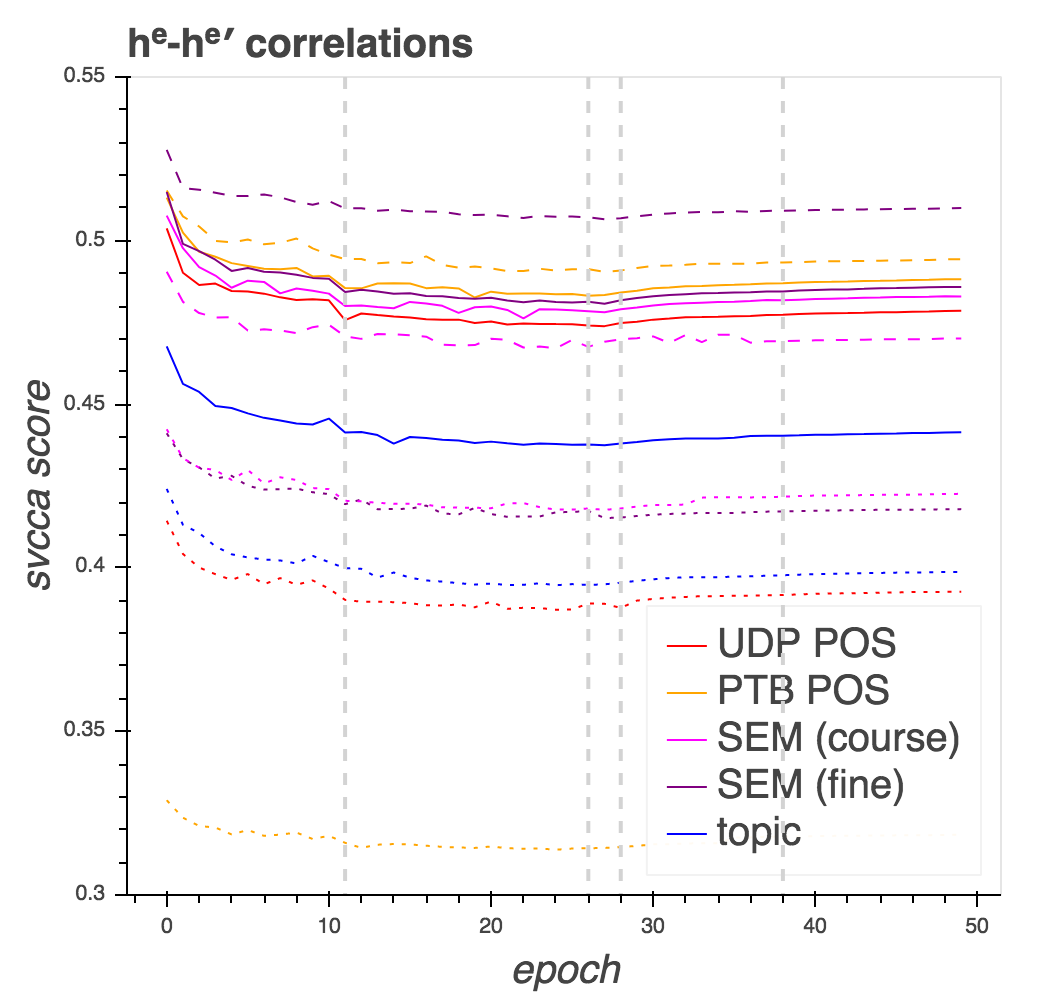}
\caption{SVCCA correlation scores between LM and $y_{t+1}$ tag predictor. Dotted lines use models trained on randomly shuffled the data. Dashed lines use GMB domain test data.}
\label{fig:domain_cca}
\end{figure}

\begin{figure}[hb]
\includegraphics[width=0.5\textwidth]{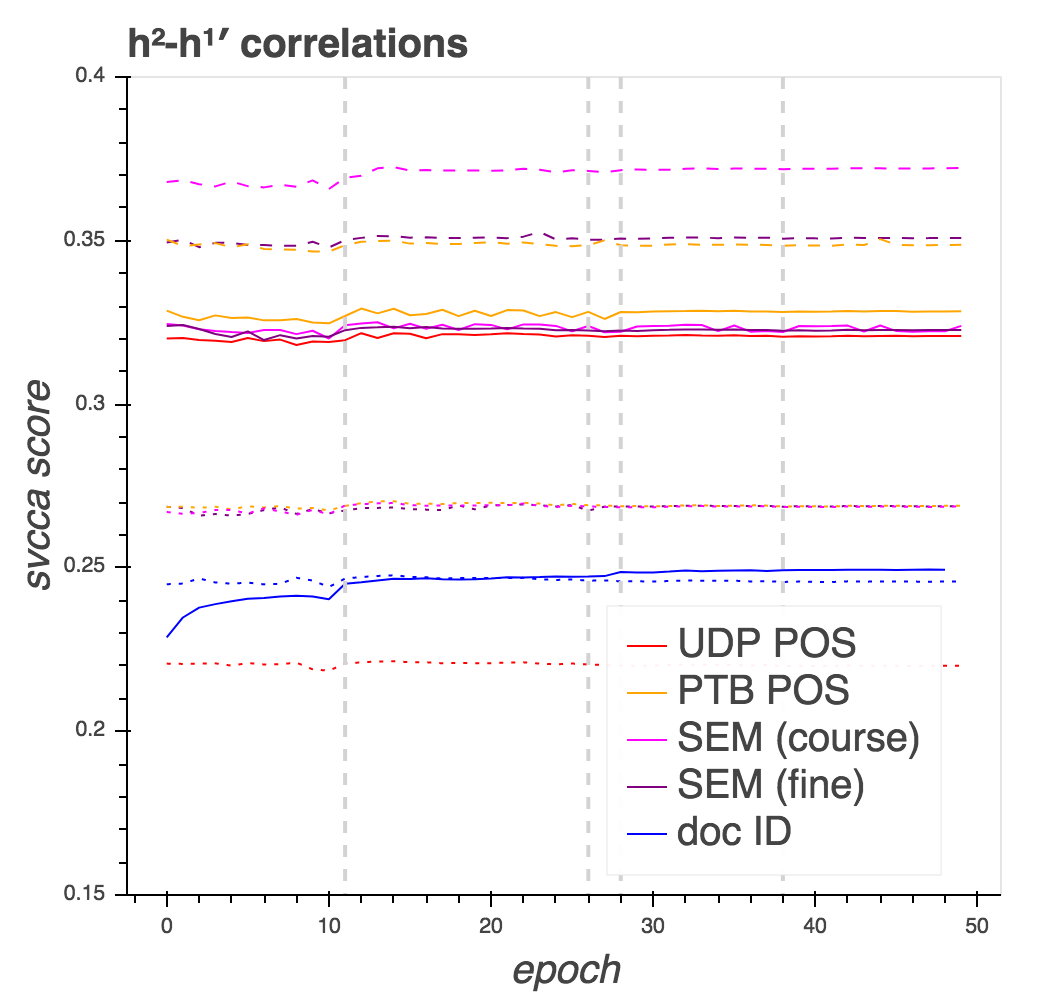}
\includegraphics[width=0.5\textwidth]{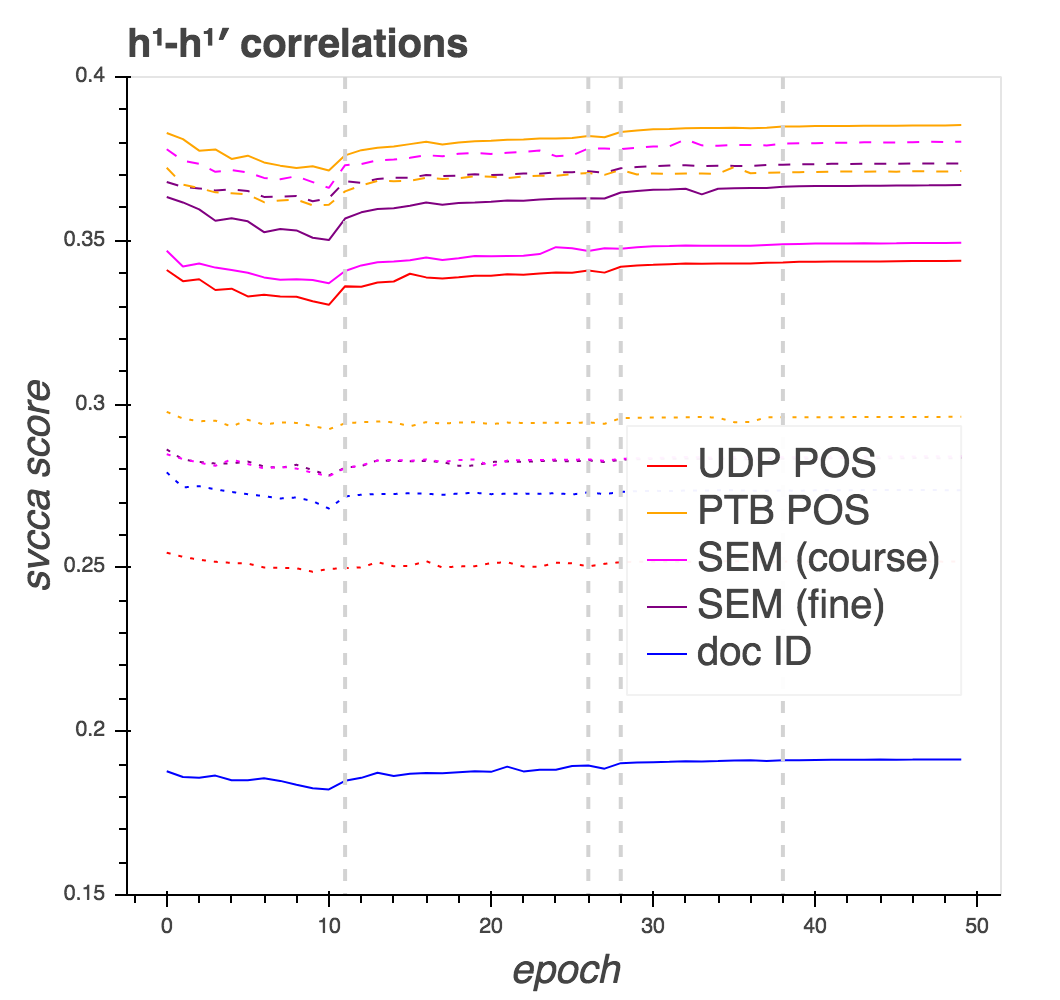}
\includegraphics[width=0.5\textwidth]{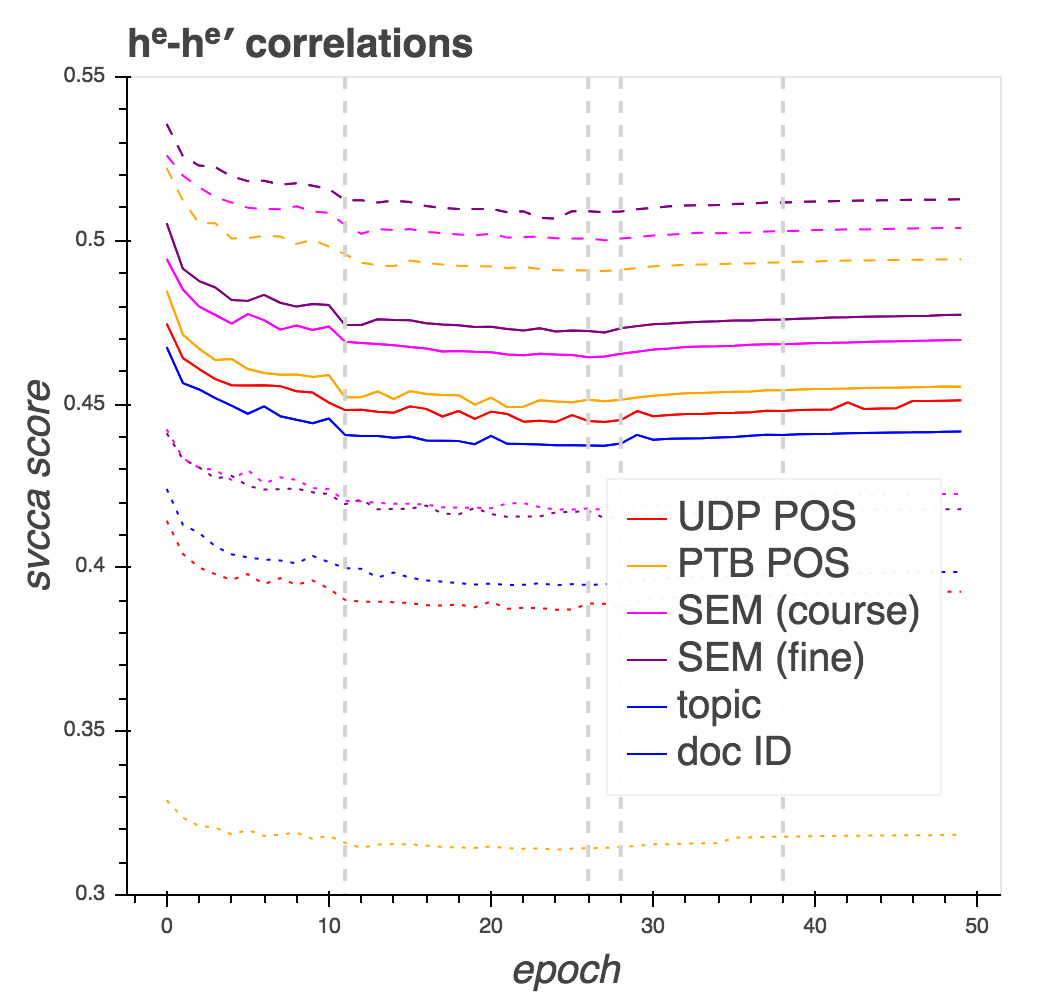}
\caption{SVCCA correlation scores between LM and $y_{t}$ tagger. Dotted lines use models trained on randomly shuffled the data. Dashed lines use GMB domain test data.}
\label{fig:domain_cca_input}
\end{figure}

Because SEM tags and PTB POS tags were both trained on the GMB corpus, we present the SVCCA similarities on an in-domain GMB test corpus as well as the Wikipedia test corpus used elsewhere in the paper. The results are in Figures~\ref{fig:domain_cca}-\ref{fig:domain_cca_input}. In general correlations are higher using the original tagging domain, but not enough to contradict our earlier analysis. The shapes of the curves remain similar.

\end{document}